\documentclass{article}

\PassOptionsToPackage{numbers,sort&compress}{natbib}
\usepackage{iclr2026_conference,times}
\setcitestyle{numbers,square,comma}

\usepackage{hyperref}
\usepackage{url}
\usepackage{pifont}
\usepackage{todonotes}
\usepackage{xcolor}
\usepackage{tikz}
\usepackage{comment}
\usepackage{subcaption}
\usepackage{placeins}
\usepackage{nameref}

\usepackage{amsmath,amsfonts,bm}

\def\eqref#1{equation~\ref{#1}}

\def\1{\bm{1}}

\DeclareMathAlphabet{\mathsfit}{\encodingdefault}{\sfdefault}{m}{sl}
\SetMathAlphabet{\mathsfit}{bold}{\encodingdefault}{\sfdefault}{bx}{n}

\definecolor{token1}{RGB}{199,178,255}  %
\definecolor{token2}{RGB}{198,255,198}  %
\definecolor{token3}{RGB}{255,218,185}  %
\definecolor{token4}{RGB}{255,182,193}  %
\definecolor{token5}{RGB}{173,216,230}  %

\newcommand{\tokenbox}[2]{%
    \tikz[baseline=(X.base)]{
        \node[
            fill=#1, 
            inner sep=0pt,        %
            inner xsep=0.0em,    %
            inner ysep=0.1em,    %
            outer sep=0pt, 
            rounded corners=0pt, 
            font=\LARGE,
        ] (X) {\strut #2};
    }%
}

\newcommand{\tokenboxinline}[2]{%
    \tikz[baseline=(X.base)]{
        \node[
            fill=#1, 
            inner sep=0pt,
            inner xsep=0.0em,
            inner ysep=0.1em,
            outer sep=0pt, 
            rounded corners=0pt, 
            font=\small,
        ] (X) {\vphantom{Ay}#2};  %
    }%
}

\usepackage{tikz}
\usetikzlibrary{calc,backgrounds}

\title{
    \tokenbox{token1}{how}%
    \tokenbox{token2}{ long}%
    \tokenbox{token3}{ is}%
    \tokenbox{token4}{ a}%
    \tokenbox{token5}{ pie}%
    \tokenbox{token1}{ce}%
    \tokenbox{token2}{ of}%
    \tokenbox{token3}{ string}%
    \tokenbox{token4}{?}\\
    \LARGE a brief empirical analysis of tokenizers
}

\author{
  \textbf{Jonathan Roberts\textsuperscript{\textcolor{teal}{$\psi$}}} \quad\quad
  \textbf{Kai Han\textsuperscript{\textcolor{orange}{$\pi$}}} \quad\quad
  \textbf{Samuel Albanie}
\\
  \textsuperscript{\textcolor{teal}{$\psi$}}University of Cambridge \quad\quad
  \textsuperscript{\textcolor{orange}{$\pi$}}The University of Hong Kong
  }

\iclrfinalcopy
\begin{document}

\maketitle

\begin{abstract}

Frontier LLMs are increasingly utilised across academia, society and industry. A commonly used unit for comparing models, their inputs and outputs, and estimating inference pricing is the token. In general, tokens are used as a stable currency, assumed to be broadly consistent across  tokenizers and contexts, enabling direct comparisons. However, %
tokenization varies significantly across models and domains of text, making na\"{\i}ve interpretation of token counts problematic. We quantify this variation by providing a comprehensive empirical analysis of tokenization, exploring the compression of sequences to tokens across different distributions of textual data. Our analysis challenges commonly held heuristics about token lengths, finding them to be overly simplistic. We hope the insights of our study add clarity and intuition toward tokenization in contemporary LLMs.

\end{abstract}

\section{Introduction}
\label{sec:intro}

Large language models (LLMs) are ubiquitous across contemporary AI research. As model capabilities continue to improve%
, LLMs are capturing attention more broadly, in society and industry. Frontier models follow instructions with sufficient consistency to robustly use tools, enabling them to act as agents capable of performing longer horizon tasks \citep{kwa2025measuring} and economically valuable activity such as software engineering \cite{miserendino2025swe, xu2024theagentcompany}, scientific research \cite{schmidgall2025agent}, and web-based economic tasks \cite{liu2025econwebarena}. %

Fundamental to LLMs is an often overlooked process: \textit{tokenization}. Tokenization describes the learned conversion between text characters and discrete ``tokens'' represented by unique numeric IDs.
These token IDs correspond to items in a vocabulary---a database of all possible tokens a model can interpret and generate. When autoregressively generating new tokens an LLM samples from a probability distribution over this vocabulary. Tokenization is necessary as it enables the transformation of human-readable text characters into a numeric model-readable format. Each token ID corresponds to an embedding---a high-dimensional numeric representation that aims to capture its semantic meaning.%

\begin{figure}[t]
\small{
\centering
\begin{minipage}[t]{0.48\textwidth}
\begin{tabbing}
\hspace{2.3cm} \= \kill
\textit{\textbf{Llama 1,2}} \>\tokenboxinline{token1}{ant}%
\tokenboxinline{token2}{id}%
\tokenboxinline{token3}{is}%
\tokenboxinline{token4}{est}%
\tokenboxinline{token5}{ab}%
\tokenboxinline{token1}{lish}%
\tokenboxinline{token2}{ment}%
\tokenboxinline{token3}{arian}%
\tokenboxinline{token4}{ism}%
\hspace{0.3cm}\textcolor{purple}{\textbf{9}}\\
\textit{\textbf{Mistral}} \>\tokenboxinline{token1}{ant}%
\tokenboxinline{token2}{id}%
\tokenboxinline{token3}{is}%
\tokenboxinline{token4}{est}%
\tokenboxinline{token5}{ablish}%
\tokenboxinline{token1}{ment}%
\tokenboxinline{token2}{arian}%
\tokenboxinline{token3}{ism}%
\hspace{0.3cm}\textcolor{purple}{\textbf{8}}\\
\textit{\textbf{Claude 4.5 (est.)}} \>\tokenboxinline{token1}{ant}%
\tokenboxinline{token2}{id}%
\tokenboxinline{token3}{is}%
\tokenboxinline{token4}{establishment}%
\tokenboxinline{token5}{ar}%
\tokenboxinline{token1}{ian}%
\tokenboxinline{token2}{is}%
\tokenboxinline{token3}{m}%
\hspace{0.3cm}\textcolor{purple}{\textbf{8}}\\
\textit{\textbf{Mistral Tekken}} \>\tokenboxinline{token1}{ant}%
\tokenboxinline{token2}{idis}%
\tokenboxinline{token3}{est}%
\tokenboxinline{token4}{abl}%
\tokenboxinline{token5}{ishment}%
\tokenboxinline{token1}{arian}%
\tokenboxinline{token2}{ism}%
\hspace{0.3cm}\textcolor{purple}{\textbf{7}}\\
\textit{\textbf{GPT 5}} \>\tokenboxinline{token1}{ant}%
\tokenboxinline{token2}{idis}%
\tokenboxinline{token3}{est}%
\tokenboxinline{token4}{ablishment}%
\tokenboxinline{token5}{arian}%
\tokenboxinline{token1}{ism}%
\hspace{0.3cm}\textcolor{purple}{\textbf{6}}\\
\textit{\textbf{OLMo}} \>\tokenboxinline{token1}{ant}%
\tokenboxinline{token2}{idis}%
\tokenboxinline{token3}{establ}%
\tokenboxinline{token4}{ishment}%
\tokenboxinline{token5}{arian}%
\tokenboxinline{token1}{ism}%
\hspace{0.3cm}\textcolor{purple}{\textbf{6}}\\
\textit{\textbf{Qwen}} \>\tokenboxinline{token1}{ant}%
\tokenboxinline{token2}{idis}%
\tokenboxinline{token3}{establish}%
\tokenboxinline{token4}{ment}%
\tokenboxinline{token5}{arian}%
\tokenboxinline{token1}{ism}%
\hspace{0.3cm}\textcolor{purple}{\textbf{6}}\\
\textit{\textbf{DeepSeek R1/V3}} \>\tokenboxinline{token1}{ant}%
\tokenboxinline{token2}{idis}%
\tokenboxinline{token3}{establish}%
\tokenboxinline{token4}{ment}%
\tokenboxinline{token5}{arianism}%
\hspace{0.3cm}\textcolor{purple}{\textbf{5}}\\
\textit{\textbf{Grok}} \>\tokenboxinline{token1}{ant}%
\tokenboxinline{token2}{idis}%
\tokenboxinline{token3}{establishment}%
\tokenboxinline{token4}{arian}%
\tokenboxinline{token5}{ism}%
\hspace{0.3cm}\textcolor{purple}{\textbf{5}}\\
\textit{\textbf{T5}} \>\tokenboxinline{token1}{anti}%
\tokenboxinline{token2}{d}%
\tokenboxinline{token3}{is}%
\tokenboxinline{token4}{est}%
\tokenboxinline{token5}{abl}%
\tokenboxinline{token1}{ish}%
\tokenboxinline{token2}{ment}%
\tokenboxinline{token3}{aria}%
\tokenboxinline{token4}{nism}%
\hspace{0.3cm}\textcolor{purple}{\textbf{9}}\\
\textit{\textbf{BERT}} \>\tokenboxinline{token1}{anti}%
\tokenboxinline{token2}{dis}%
\tokenboxinline{token3}{est}%
\tokenboxinline{token4}{ab}%
\tokenboxinline{token5}{lish}%
\tokenboxinline{token1}{ment}%
\tokenboxinline{token2}{arian}%
\tokenboxinline{token3}{ism}%
\hspace{0.3cm}\textcolor{purple}{\textbf{8}}\\
\textit{\textbf{Gemini}} \>\tokenboxinline{token1}{anti}%
\tokenboxinline{token2}{dis}%
\tokenboxinline{token3}{establishment}%
\tokenboxinline{token4}{arian}%
\tokenboxinline{token5}{ism}%
\hspace{0.3cm}\textcolor{purple}{\textbf{5}}\\
\textit{\textbf{Fuyu}} \>\tokenboxinline{token1}{anti}%
\tokenboxinline{token2}{dis}%
\tokenboxinline{token3}{establishment}%
\tokenboxinline{token4}{arianism}%
\hspace{0.3cm}\textcolor{purple}{\textbf{4}}\\
\end{tabbing}
\end{minipage}%
\hfill%
\begin{minipage}[t]{0.48\textwidth}
\begin{tabbing}
\hspace{2.3cm} \= \kill
\textit{\textbf{Llama 1,2}} \>\tokenboxinline{token1}{hum}%
\tokenboxinline{token2}{u}%
\tokenboxinline{token3}{hum}%
\tokenboxinline{token4}{un}%
\tokenboxinline{token5}{uk}%
\tokenboxinline{token1}{un}%
\tokenboxinline{token2}{u}%
\tokenboxinline{token3}{ku}%
\tokenboxinline{token4}{ā}%
\tokenboxinline{token5}{p}%
\tokenboxinline{token1}{ua}%
\tokenboxinline{token2}{'}%
\tokenboxinline{token3}{a}%
\hspace{0.3cm}\textcolor{purple}{\textbf{13}}\\
\textit{\textbf{T5}} \>\tokenboxinline{token1}{hum}%
\tokenboxinline{token2}{u}%
\tokenboxinline{token3}{hum}%
\tokenboxinline{token4}{un}%
\tokenboxinline{token5}{uk}%
\tokenboxinline{token1}{un}%
\tokenboxinline{token2}{uk}%
\tokenboxinline{token3}{u}%
\tokenboxinline{token4}{ā}%
\tokenboxinline{token5}{pu}%
\tokenboxinline{token1}{a}%
\tokenboxinline{token2}{'}%
\tokenboxinline{token3}{a}%
\hspace{0.3cm}\textcolor{purple}{\textbf{13}}\\
\textit{\textbf{DeepSeek V2}} \>\tokenboxinline{token1}{hum}%
\tokenboxinline{token2}{u}%
\tokenboxinline{token3}{hum}%
\tokenboxinline{token4}{un}%
\tokenboxinline{token5}{uk}%
\tokenboxinline{token1}{un}%
\tokenboxinline{token2}{uku}%
\tokenboxinline{token3}{ā}%
\tokenboxinline{token4}{p}%
\tokenboxinline{token5}{ua}%
\tokenboxinline{token1}{'}%
\tokenboxinline{token2}{a}%
\hspace{0.3cm}\textcolor{purple}{\textbf{12}}\\
\textit{\textbf{Fuyu}} \>\tokenboxinline{token1}{hum}%
\tokenboxinline{token2}{u}%
\tokenboxinline{token3}{hum}%
\tokenboxinline{token4}{unu}%
\tokenboxinline{token5}{kun}%
\tokenboxinline{token1}{uku}%
\tokenboxinline{token2}{āp}%
\tokenboxinline{token3}{ua}%
\tokenboxinline{token4}{'}%
\tokenboxinline{token5}{a}%
\hspace{0.3cm}\textcolor{purple}{\textbf{10}}\\
\textit{\textbf{Yi}} \>\tokenboxinline{token1}{hum}%
\tokenboxinline{token2}{uh}%
\tokenboxinline{token3}{um}%
\tokenboxinline{token4}{un}%
\tokenboxinline{token5}{uk}%
\tokenboxinline{token1}{un}%
\tokenboxinline{token2}{uk}%
\tokenboxinline{token3}{u}%
\tokenboxinline{token4}{ā}%
\tokenboxinline{token5}{p}%
\tokenboxinline{token1}{ua}%
\tokenboxinline{token2}{'}%
\tokenboxinline{token3}{a}%
\hspace{0.3cm}\textcolor{purple}{\textbf{13}}\\
\textit{\textbf{Mistral}} \>\tokenboxinline{token1}{hum}%
\tokenboxinline{token2}{uh}%
\tokenboxinline{token3}{um}%
\tokenboxinline{token4}{un}%
\tokenboxinline{token5}{uk}%
\tokenboxinline{token1}{un}%
\tokenboxinline{token2}{uk}%
\tokenboxinline{token3}{u}%
\tokenboxinline{token4}{ā}%
\tokenboxinline{token5}{pu}%
\tokenboxinline{token1}{a}%
\tokenboxinline{token2}{'}%
\tokenboxinline{token3}{a}%
\hspace{0.3cm}\textcolor{purple}{\textbf{13}}\\
\textit{\textbf{BERT}} \>\tokenboxinline{token1}{hum}%
\tokenboxinline{token2}{uh}%
\tokenboxinline{token3}{um}%
\tokenboxinline{token4}{un}%
\tokenboxinline{token5}{uk}%
\tokenboxinline{token1}{un}%
\tokenboxinline{token2}{uk}%
\tokenboxinline{token3}{ua}%
\tokenboxinline{token4}{pu}%
\tokenboxinline{token5}{a}%
\tokenboxinline{token1}{'}%
\tokenboxinline{token2}{a}%
\hspace{0.3cm}\textcolor{purple}{\textbf{12}}\\
\textit{\textbf{Jamba}} \>\tokenboxinline{token1}{hum}%
\tokenboxinline{token2}{uh}%
\tokenboxinline{token3}{um}%
\tokenboxinline{token4}{un}%
\tokenboxinline{token5}{uk}%
\tokenboxinline{token1}{un}%
\tokenboxinline{token2}{uku}%
\tokenboxinline{token3}{ā}%
\tokenboxinline{token4}{p}%
\tokenboxinline{token5}{ua}%
\tokenboxinline{token1}{'}%
\tokenboxinline{token2}{a}%
\hspace{0.3cm}\textcolor{purple}{\textbf{12}}\\
\textit{\textbf{Qwen}} \>\tokenboxinline{token1}{hum}%
\tokenboxinline{token2}{uh}%
\tokenboxinline{token3}{um}%
\tokenboxinline{token4}{un}%
\tokenboxinline{token5}{uk}%
\tokenboxinline{token1}{un}%
\tokenboxinline{token2}{uku}%
\tokenboxinline{token3}{ā}%
\tokenboxinline{token4}{p}%
\tokenboxinline{token5}{ua}%
\tokenboxinline{token1}{'a}%
\hspace{0.3cm}\textcolor{purple}{\textbf{11}}\\
\textit{\textbf{Kimi K2}} \>\tokenboxinline{token1}{hum}%
\tokenboxinline{token2}{uh}%
\tokenboxinline{token3}{um}%
\tokenboxinline{token4}{un}%
\tokenboxinline{token5}{uk}%
\tokenboxinline{token1}{un}%
\tokenboxinline{token2}{uku}%
\tokenboxinline{token3}{ā}%
\tokenboxinline{token4}{pu}%
\tokenboxinline{token5}{a}%
\tokenboxinline{token1}{'a}%
\hspace{0.3cm}\textcolor{purple}{\textbf{11}}\\
\textit{\textbf{Gemini}} \>\tokenboxinline{token1}{hum}%
\tokenboxinline{token2}{uh}%
\tokenboxinline{token3}{um}%
\tokenboxinline{token4}{un}%
\tokenboxinline{token5}{ukun}%
\tokenboxinline{token1}{uku}%
\tokenboxinline{token2}{ā}%
\tokenboxinline{token3}{p}%
\tokenboxinline{token4}{ua}%
\tokenboxinline{token5}{'}%
\tokenboxinline{token1}{a}%
\hspace{0.3cm}\textcolor{purple}{\textbf{11}}\\
\textit{\textbf{DeepSeek R1/V3}} \>\tokenboxinline{token1}{hum}%
\tokenboxinline{token2}{uh}%
\tokenboxinline{token3}{um}%
\tokenboxinline{token4}{un}%
\tokenboxinline{token5}{ukun}%
\tokenboxinline{token1}{uku}%
\tokenboxinline{token2}{ā}%
\tokenboxinline{token3}{p}%
\tokenboxinline{token4}{ua}%
\tokenboxinline{token5}{'a}%
\hspace{0.3cm}\textcolor{purple}{\textbf{10}}\\
\textit{\textbf{Grok}} \>\tokenboxinline{token1}{hum}%
\tokenboxinline{token2}{uh}%
\tokenboxinline{token3}{um}%
\tokenboxinline{token4}{un}%
\tokenboxinline{token5}{ukun}%
\tokenboxinline{token1}{uku}%
\tokenboxinline{token2}{āp}%
\tokenboxinline{token3}{ua}%
\tokenboxinline{token4}{'a}%
\hspace{0.3cm}\textcolor{purple}{\textbf{9}}\\
\end{tabbing}
\end{minipage}

\vspace{-3mm}

\begin{tabbing}
\hspace{4.5cm} \= \kill
\textit{\textbf{Llama 1,2}} \>\tokenboxinline{token1}{A}%
\tokenboxinline{token2}{ fo}%
\tokenboxinline{token3}{x}%
\tokenboxinline{token4}{ knows}%
\tokenboxinline{token5}{ many}%
\tokenboxinline{token1}{ things}%
\tokenboxinline{token2}{,}%
\tokenboxinline{token3}{ but}%
\tokenboxinline{token4}{ a}%
\tokenboxinline{token5}{ h}%
\tokenboxinline{token1}{edge}%
\tokenboxinline{token2}{h}%
\tokenboxinline{token3}{og}%
\tokenboxinline{token4}{ knows}%
\tokenboxinline{token5}{ one}%
\tokenboxinline{token1}{ big}%
\tokenboxinline{token2}{ thing}%
\tokenboxinline{token3}{.}%
$^*$
\hspace{0.07cm}\textcolor{purple}{\textbf{18}}
\\
\textit{\textbf{Llama 3}} \>\tokenboxinline{token1}{A}%
\tokenboxinline{token2}{ fox}%
\tokenboxinline{token3}{ knows}%
\tokenboxinline{token4}{ many}%
\tokenboxinline{token5}{ things}%
\tokenboxinline{token1}{,}%
\tokenboxinline{token2}{ but}%
\tokenboxinline{token3}{ a}%
\tokenboxinline{token4}{ hedge}%
\tokenboxinline{token5}{hog}%
\tokenboxinline{token1}{ knows}%
\tokenboxinline{token2}{ one}%
\tokenboxinline{token3}{ big}%
\tokenboxinline{token4}{ thing}%
\tokenboxinline{token5}{.}%
\hspace{0.3cm}\textcolor{purple}{\textbf{15}}
\\
\textit{\textbf{Llama 4}} \>\tokenboxinline{token1}{A}%
\tokenboxinline{token2}{ fox}%
\tokenboxinline{token3}{ knows}%
\tokenboxinline{token4}{ many}%
\tokenboxinline{token5}{ things}%
\tokenboxinline{token1}{,}%
\tokenboxinline{token2}{ but}%
\tokenboxinline{token3}{ a}%
\tokenboxinline{token4}{ hed}%
\tokenboxinline{token5}{geh}%
\tokenboxinline{token1}{og}%
\tokenboxinline{token2}{ knows}%
\tokenboxinline{token3}{ one}%
\tokenboxinline{token4}{ big}%
\tokenboxinline{token5}{ thing}%
\tokenboxinline{token1}{.}%
\hspace{0.3cm}\textcolor{purple}{\textbf{16}}
\\
\end{tabbing}
\vspace{-6mm}
\begin{tabbing}
\hspace{4.5cm} \= \kill
\textit{\textbf{Mistral}} \>\tokenboxinline{token1}{A}%
\tokenboxinline{token2}{ f}%
\tokenboxinline{token3}{ox}%
\tokenboxinline{token4}{ knows}%
\tokenboxinline{token5}{ many}%
\tokenboxinline{token1}{ things}%
\tokenboxinline{token2}{,}%
\tokenboxinline{token3}{ but}%
\tokenboxinline{token4}{ a}%
\tokenboxinline{token5}{ hed}%
\tokenboxinline{token1}{ge}%
\tokenboxinline{token2}{h}%
\tokenboxinline{token3}{og}%
\tokenboxinline{token4}{ knows}%
\tokenboxinline{token5}{ one}%
\tokenboxinline{token1}{ big}%
\tokenboxinline{token2}{ thing}%
\tokenboxinline{token3}{.}%
\hspace{0.3cm}\textcolor{purple}{\textbf{18}}
\\
\textit{\textbf{Mistral Tekken}} \>\tokenboxinline{token1}{A}%
\tokenboxinline{token2}{  fox}%
\tokenboxinline{token3}{  knows}%
\tokenboxinline{token4}{  many}%
\tokenboxinline{token5}{  things}%
\tokenboxinline{token1}{,}%
\tokenboxinline{token2}{  but}%
\tokenboxinline{token3}{  a}%
\tokenboxinline{token4}{  hed}%
\tokenboxinline{token5}{geh}%
\tokenboxinline{token1}{og}%
\tokenboxinline{token2}{  knows}%
\tokenboxinline{token3}{  one}%
\tokenboxinline{token4}{  big}%
\tokenboxinline{token5}{  thing}%
\tokenboxinline{token1}{.}%
\hspace{0.3cm}\textcolor{purple}{\textbf{16}}
\\
\end{tabbing}
\vspace{-6mm}
\begin{tabbing}
\hspace{4.5cm} \= \kill
\textit{\textbf{GPT 3.5/4}} \>\tokenboxinline{token1}{A}%
\tokenboxinline{token2}{  fox}%
\tokenboxinline{token3}{  knows}%
\tokenboxinline{token4}{  many}%
\tokenboxinline{token5}{  things}%
\tokenboxinline{token1}{,}%
\tokenboxinline{token2}{  but}%
\tokenboxinline{token3}{  a}%
\tokenboxinline{token4}{  hedge}%
\tokenboxinline{token5}{hog}%
\tokenboxinline{token1}{  knows}%
\tokenboxinline{token2}{  one}%
\tokenboxinline{token3}{  big}%
\tokenboxinline{token4}{  thing}%
\tokenboxinline{token5}{.}%
\hspace{0.3cm}\textcolor{purple}{\textbf{15}}
\\
\textit{\textbf{GPT 4o/4.1/4.5/5; oss; o-series}} \>\tokenboxinline{token1}{A}%
\tokenboxinline{token2}{  fox}%
\tokenboxinline{token3}{  knows}%
\tokenboxinline{token4}{  many}%
\tokenboxinline{token5}{  things}%
\tokenboxinline{token1}{,}%
\tokenboxinline{token2}{  but}%
\tokenboxinline{token3}{  a}%
\tokenboxinline{token4}{  hed}%
\tokenboxinline{token5}{geh}%
\tokenboxinline{token1}{og}%
\tokenboxinline{token2}{  knows}%
\tokenboxinline{token3}{  one}%
\tokenboxinline{token4}{  big}%
\tokenboxinline{token5}{  thing}%
\tokenboxinline{token1}{.}%
\hspace{0.3cm}\textcolor{purple}{\textbf{16}}
\end{tabbing}

\vspace{-1\baselineskip}
\caption{\textbf{Tokenization schemes vary significantly}. Token boundaries are represented as shaded colours and token counts are shown in \textbf{\textcolor{purple}{purple}}. \textit{Antidisestablishmentarianism} is tokenized many different ways, using %
between \textcolor{purple}{\textbf{4}} (Fuyu) and \textcolor{purple}{\textbf{9}} tokens (early Llama models). Variation in tokenization is also observable for words containing fewer unique letters and more vowels,  such as the Hawaiian word \textit{humuhumunukunukuāpua'a} (Reef Triggerfish). Small changes to tokenizers within model families result in different tokenization, even in relatively simple sentences. $^*$From \cite{berlin2013hedgehog}.}
\label{fig:example_tokenization}
}
\end{figure}

The token %
is a near-universally used unit of text sequences for describing numerous language modelling metrics, including sequence lengths, context limits, pricing, latency, and sizes of training corpora. %
Token counts are utilised to directly compare models, inference costs and provider platforms. Although different model series are known to implement bespoke tokenizers, and therefore tokenize text differently, tokens are used as though they are a consistent unit with differences that are trivial or averaged out over sufficient samplings or length of text. This assumption is evident in the widespread use of general heuristics for token lengths (such as one token is roughly 4 characters or 0.75 words (\textit{e.g.}, \cite{OpenAI_Tokenizer, Gemini_Tokenizer})); or the comparison of API endpoint performance using number of tokens per second or usage metered as number of dollars per token.

However, as we demonstrate, token counts are not a stable unit of length: counts vary non-trivially for different domains of text or tokenizer. Therefore, when used na\"{\i}vely, token counts provide an inadequate basis of comparison. In Fig. \ref{fig:example_tokenization}, we show clear variations in token boundaries and the number of tokens used by different tokenization schemes. Among frontier LLMs, a word like \textit{antidisestablishmentarianism} is tokenized by Claude into nearly twice as many tokens (9) as Gemini or Grok (5). As our experimentation demonstrates, these differences in tokenization can result in significant differences in token counts over longer sequences.

As LLM-based systems become more prevalent, there is an emerging need for clarity around tokenization. Simply comparing sequence lengths and context limits based on model-specific tokenization is insufficient and misleading. Moreover, when accessed via an API (as is the case for most frontier LLMs), usage is metered by the token: accurate accounting requires a clear understanding of how a text sequence is mapped to tokens for a given text sequence.

Prior studies have explored tokenization for specific domains \cite{roberts2025needle}, languages \cite{ahia2023all} or strategies \cite{saleva2023changes}, painting a valuable though incomplete picture. We extend these works and offer a comprehensive empirical analysis of tokenizer efficiencies across models and domains. We provide robust quantification of domain-specific tokenizer character compression ratios and estimates of word compression ratios as a function of word frequency and distribution. Building on these empirical findings, we examine model context limits, suggesting domain-specific character counts or model-agnostic token counts offer more directly comparable context limits. Our study is intentionally brief and focused, with clear goals of adding clarity to tokenization, quantifying differences in tokenization between models and domains, and imparting intuition on how to interpret token-based metrics.

\section{Tokenization}

During inference, an input text string is encoded into a sequence of token IDs by the tokenizer. These token IDs map to embeddings, which the LLM ingests along with positional information as input and iteratively generates a sequence of output token IDs. Finally, the tokenizer decodes this output into a human-readable text string.
The exact nature of the mapping between formats and the level of compression between text characters and tokens is a trade-off balancing \textit{efficiency} and \textit{meaning}. Tokenizing at the character-level ensures a relatively small vocabulary, constrained by the number of unique characters, resulting in smaller embedding/output matrices at the expense of longer sequences and semantic meaning (information density) per token (`c' carries less meaning than `car'). On the other hand, tokenizing at the word-level (or longer) results in fewer tokens that capture more semantic meaning at the expense of a larger vocabulary%
; this approach is subject to out-of-vocabulary issues. To compromise these conflicting objectives, tokenizers in LLMs are typically trained to tokenize at the subword level.

Tokenization typically consists of two phases: \textit{training} and \textit{segmentation}. \textbf{Training}: Several methods are commonly used to train tokenizers \cite{sennrich2016neural, schuster2012japanese, kudo2018subword, kudo2018sentencepiece}. Fundamental to all is the use of occurrence frequency of text strings within a training corpus. For methods like byte-pair encoding (BPE) \cite{sennrich2016neural} or WordPiece \cite{schuster2012japanese}, training involves iteratively merging bigrams in the corpus until the tokens defined by the merges fill the vocabulary to a desired size. \textbf{Segmentation}: Training creates a vocabulary of allowed tokens but does not directly enable tokenization---rules are needed to segment text sequences into tokens such as determining the order of tokenization and where to draw token boundaries when potential token strings overlap. For most methods, this follows the merge rules and occurrence probabilities determined during training, and thus is implicitly influenced by the structure of the training corpus.

Differences in the tokenization of a text sequence occur along two axes: the \textit{tokenizer} and the \textit{domain of the text}. The choice of tokenizer directly introduces differences as tokenizers are typically trained using slightly differing methods on different training corpora and have different segmentation algorithms, resulting in non-identical tokenization schemes. Indirect differences are introduced by the domain of the text---distributions of text containing more commonly occurring sequences of characters are tokenized at a higher rate; a general-purpose essay will likely correspond to fewer tokens than a technical paper of equal character or word length. 

\section{Domains and Tokenizers}

\paragraph{Domains.} Tab. \ref{tab:domains} outlines the 8 distinct domains we use to curate a small test corpus. We consider these domains to cover most common types of text sequences relevant to LLMs, including standard natural language, technical language, numeric sequences, and programming languages. 

\begin{table}[h]
    \centering
    \begin{tabular}{lll}
    \textbf{Domain} & \textbf{Source} & %
    \textbf{Description}\\
    \hline
    \textit{Essay} & Paul Graham essays \citep{graham_essays_web} & Tech essays in English\\
    \textit{Technical} & arXiv$^*$& 2x Oct-25 papers per arXiv category\\
    \textit{Code} & Transformers repository \citep{wolf-etal-2020-transformers} & %
    Python code (transformers 4.46)\\
    \textit{Number} & $\pi$ & %
    First 100k digits of $\pi$\\
    \textit{Emojis} & Emoji dataset \citep{abdullah_emoji_dataset_2024} & %
    $\sim$5k emojis\\
    \textit{Alphanumeric} & UUIDs & %
    Dictionary of UUID pairs\\
    \textit{Structured} & FinTabNet \citep{apoidea_fintabnet_html_2024} & Financial data extracted from tables\\
    \textit{Web} & Popular webpages & Rendered HTML from 10 websites\\
    \hline
    \end{tabular}
    \vspace{-0.2cm}
    \caption{\textbf{Text domains used in our experiments.} Unless referenced, all data was sourced by the authors; additional details can be found in the Appendix. $^*$\url{https://arxiv.org/}}
    \label{tab:domains}
\end{table}

\paragraph{Tokenizers.} We focus our experimentation on a 10-tokenizer subset of the models analysed in Fig. \ref{fig:example_tokenization}, considering these to represent a broad range of proprietary and open-weights models covering both frontier and weaker levels of capability. In many cases, tokenizers are reused across models within a series, rather than being bespoke. To increase the applicability of our results, we refer to model families used by each tokenizer and include Tab. \ref{tab:tokenizers} as a reference for specific models.

\begin{table}[h]
    \centering
    \begin{tabular}{ll}
    \textbf{Tokenizer} & \textbf{Models}\\
    \hline
    Claude \cite{claude_tokenizer} & Claude 3/3.5/4/4.5\\
    DeepSeek \cite{deepseek_tokenizer} & DeepSeek V3/R1\\
    Gemini \cite{gemini_tokenizer_doc} & Gemini 2/2.5\\
    Grok \cite{grok_tokenizer} & Grok 3/4\\
    GPT \cite{openai_tokenizer_doc} & GPT 4o/4.1/4.5/5/oss; O-Series\\
    Mistral (Tekken) \cite{mistral_tokenizer} & Mistral-Nemo/Pixtral/Ministral\\
    Llama \cite{llama4_tokenizer} & 4\\
    Reka \cite{reka_tokenizer} & Reka Flash 3/3.1\\
    Jamba \cite{jamba_tokenizer} & Jamba 1.5/1.6/1.7\\
    Qwen \cite{qwen_tokenizer} & Qwen 1.5/2/2.5/3/; QVQ/QWQ\\
    \hline
    \end{tabular}
    
    \caption{   
    \textbf{Tokenizer-model mapping}. Tokenizers are commonly shared across models. These mappings and all token counts analysis in this work are empirically-derived using the cited repositories/SDKs. Tokenizers mentioned hereafter refer to the corresponding model mappings.}
    \label{tab:tokenizers}
\end{table}

We use the term \textit{compression ratio} throughout to describe the efficiency with which characters in a text sequence are converted into tokens, \textit{i.e.}, $c = n_{chars} / n_{tokens}$. In other works \cite{rust2021good,turuta2025tokenization}, token fertility is reported as the mean number of tokens per word; to offer a more direct comparison to the 0.75 words per token heuristic, we report words per token instead. There are numerous ways of measuring the length and number of characters in a text string. In this work, we count characters as the number of Unicode code points in a given string. This approach is chosen over bytes, which are dependent on an underlying encoding (\textit{e.g.}, UTF-8), and grapheme cluster counts, which can vary with Unicode version and implementation.

\section{Experiments}

We present our core empirical analysis in 3 distinct sections, covering the following tokenization metrics: character compression (\S\ref{sec:char_comp}), word compression (\S\ref{sec:word_comp}), and sequence lengths (\S\ref{sec:seq_len}).

\subsection{Character compression}
\label{sec:char_comp}

\begin{figure}[h]
    \centering
    \includegraphics[height=0.5\linewidth, width=\linewidth]{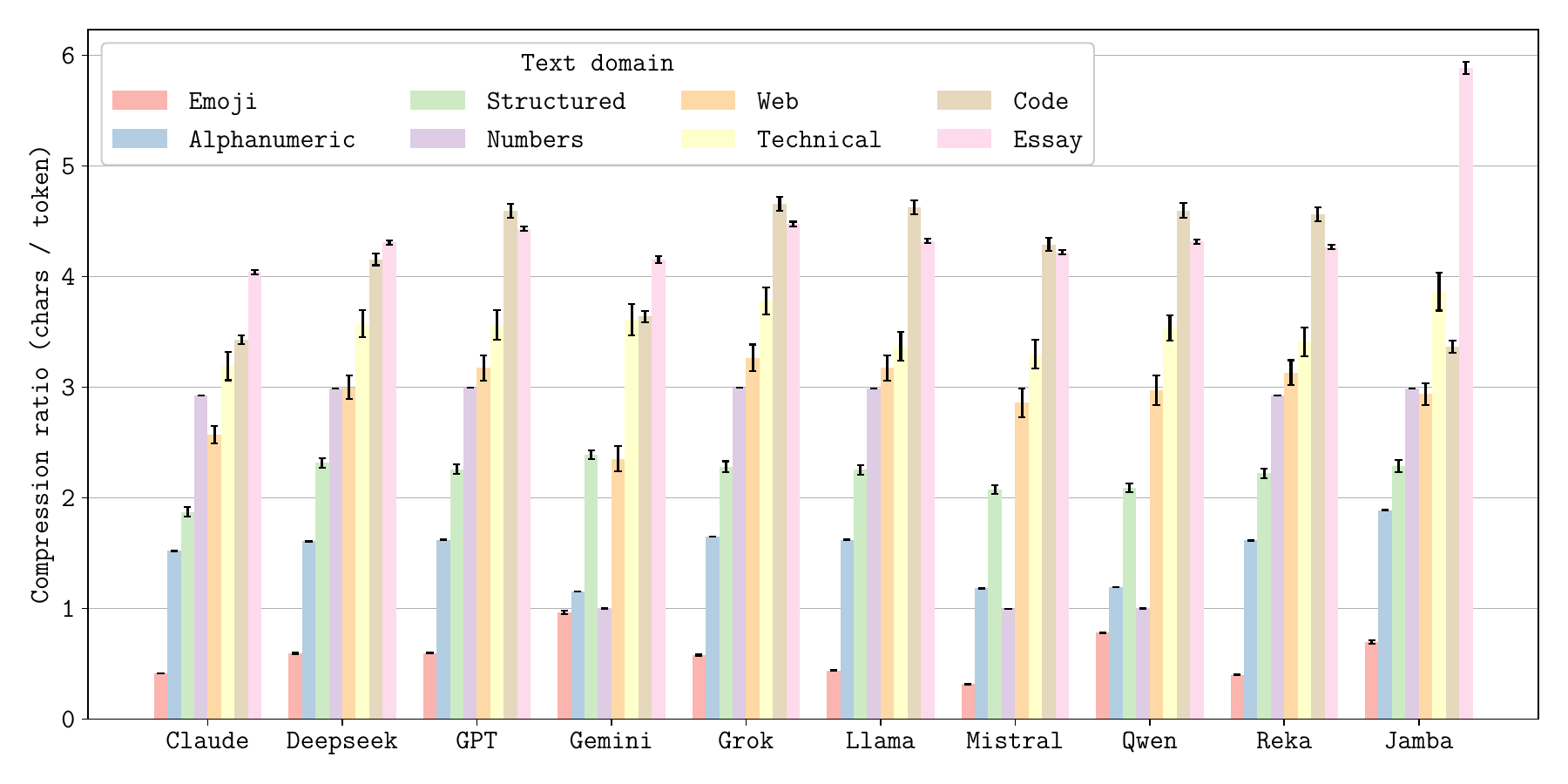}
    \vspace{-0.5cm}
    \caption{\textbf{Mean tokenizer compression ratio across different text domains}. Error bars are calculated using the standard error over 50 deterministic samplings (each $\ge$1000 characters).}
    \label{fig:compression_rates}
\end{figure}

We calculate the compression ratio of 10 LLM tokenizers by averaging over 50 deterministic samplings for each of the 8 domain text corpora described above. Each sample contained at least 1000 characters and was randomly extracted from the corpus, ensuring clean token boundaries (\textit{e.g.,} linebreaks/whitespace) at each end of the sequence. These results are displayed in Fig. \ref{fig:compression_rates}.

\textbf{Tokenizer}. As expected due to differences in tokenizer training (corpus distribution and algorithm) and segmentation, compression ratios for a given domain vary significantly across tokenizers. The more useful insight is the degree of variation---taking into account the extreme values, compression ratios differ by 100\% for the emoji and numbers domains and between $\sim$20-50\% for the others.

\textbf{Domain}. Significant variation in compression ratio is observed across domains. For most models, compression ratios vary by nearly a factor of ten between the lowest (emojis) and highest (code/essay) compression ratios. Considering just natural language domains, the compression rate for essays is approximately 25\% higher than that of technical text.

\textbf{Precision}. The size of the standard error conveys information about the breadth of the tokenizer vocabulary for text in each domain. Less structure in the text of the numeric, alphanumeric, and emoji domains results in fewer relevant tokens and a better-defined compression ratio. In the natural language domains (essay, web, technical and python code), the larger pool of different tokens produces more variation in the compression ratios across samples.

\begin{figure}
    \centering
    \includegraphics[width=\textwidth]{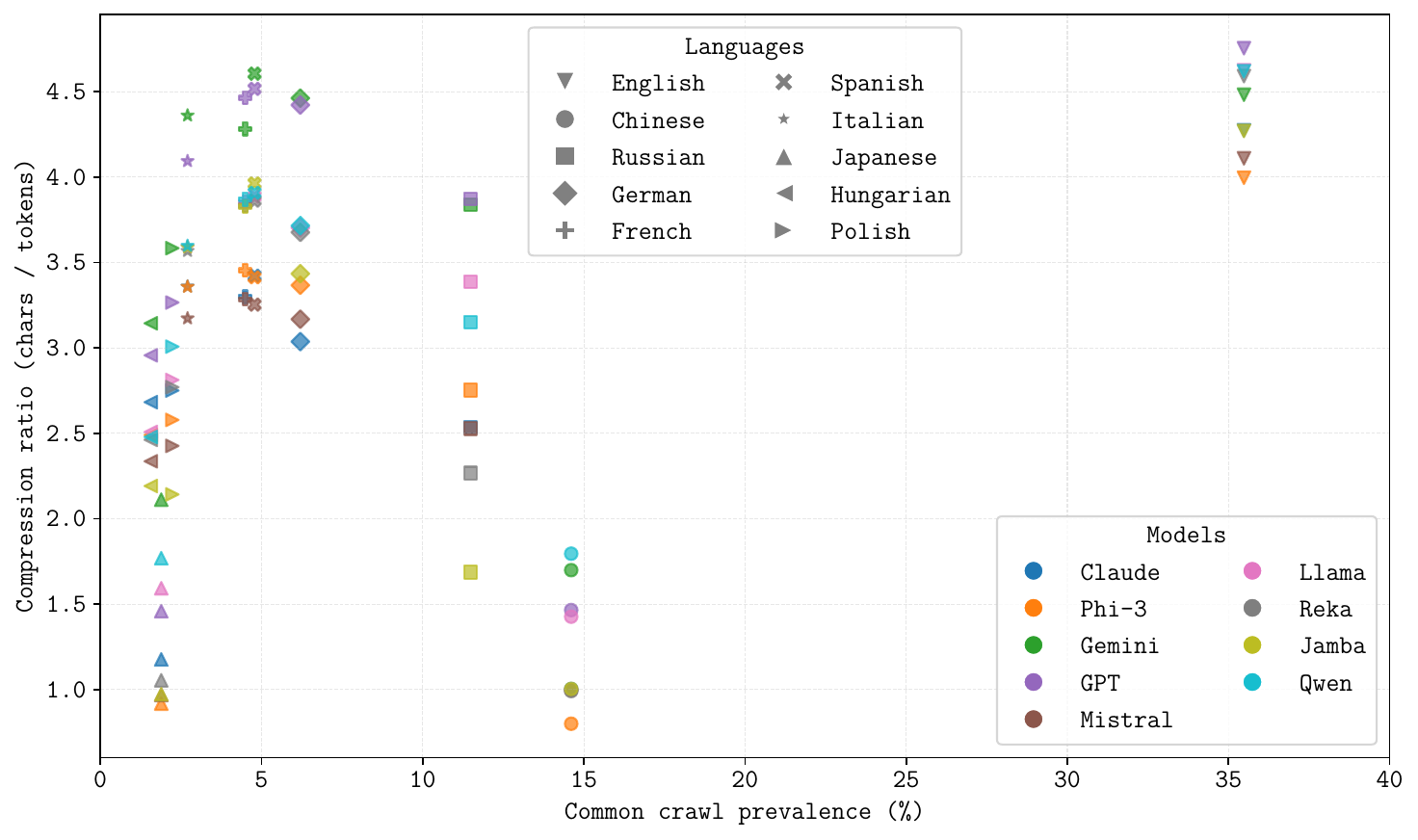}
    \vspace{-0.7cm}
    \caption{\textbf{Compression ratios of essays \cite{graham_essays_web} translated into different languages.} Compression ratios are averaged over 11 essays; common crawl prevalence is estimated from \cite{abadji-etal-2022-towards}. Where necessary, arbitrary offsets have been added to the x-coordinates to reduce overlap between languages.}
    \label{fig:language_tokenization_prev}
\end{figure}

The macro-observation from these results is that \textbf{variation in compression ratios is non-trivial}: assuming near-consistent tokenization rates across domains and models is an oversimplification. Moreover, the analysis of the compression ratio of natural language essays of different languages in Fig. \ref{fig:language_tokenization_prev} also shows differences in the spread of compression ratios across tokenizers for a given language and notable disparity between languages. There is no significant correlation between the compression ratio and the common crawl prevalence for a given language.

\subsection{Word compression}
\label{sec:word_comp}

\begin{figure}[t!]
    \centering
    \includegraphics[height=0.5\linewidth, width=\linewidth]{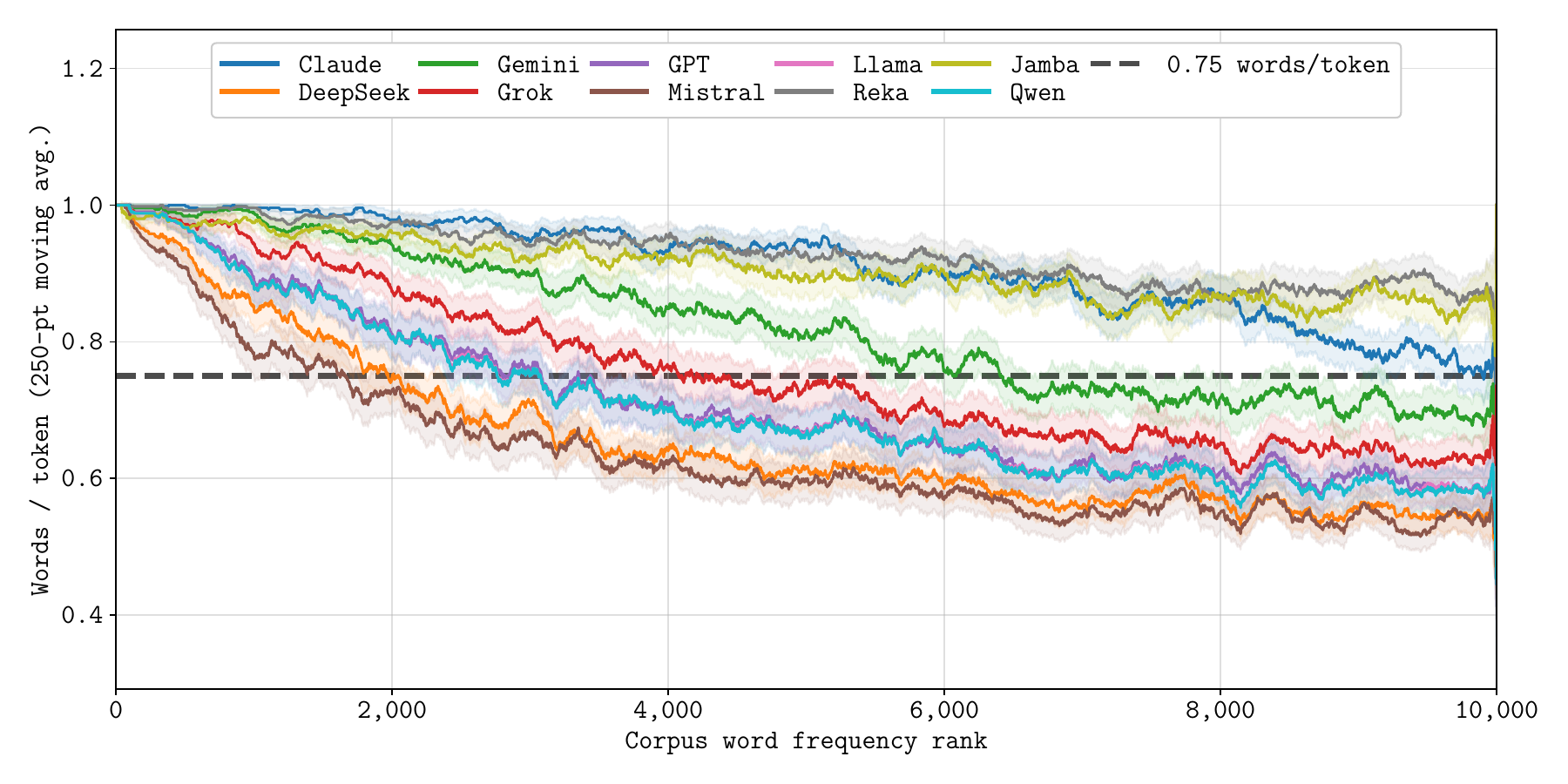}
    \vspace{-0.8cm}
    \caption{\textbf{Word to token compression variation for frequency-ranked English words}. Shaded regions show 95\% confidence intervals. Mean words/token for randomly selected English words are far lower (0.35--0.45).}
    \label{fig:token_heuristic}
\end{figure}

To investigate the common rule of thumb equating one token to approximately 0.75 words, we use a frequency-ranked set of the 10,000 most frequent English words (derived from Google's Trillion Word Corpus \cite{google10000english, brants_franz_2006_web1t5}) as a representative sample of the ``everyday" words the rule largely applies to. In Fig. \ref{fig:token_heuristic}, we plot a moving average of the words per token for the words in the corpus. Due to the influence of occurrence frequency in the tokenizer training distribution, the words per token decrease at higher frequency rank (as the words become less commonplace).

Analysis of the model lines relative to the 0.75 words/token (dashed) line challenge the rule of thumb. The rule could situationally provide a reasonable estimate for some models on text sequences that include a broad distribution of word frequency ranks. %
However, for widely used models such as Claude and Gemini, the average words/token is well above 0.75 for the first 5k most frequent words, and only just approaches 0.75 by the 10 thousandth. In these cases, the rule overestimates the number of tokens of a given text sequence. For other tokenizers, such as DeepSeek and Mistral, the rule is more of an underestimate, with the average words/token closer to 0.6 for the 10k most frequent words. The disparity between the rule and empirical values are more stark when considering randomly selected words from the English language. Average words/token for 10,000 words randomly sampled across the entire WordNet \cite{miller1995wordnet} English language lie in the range 0.35-0.45 for these models. A more detailed comparison of words/tokens for different text samples can be found in the \nameref{app}. \textbf{On balance, we find the heuristic of 0.75 words/token to be an oversimplification}.

\subsection{Context limits}
\label{sec:seq_len}

\begin{figure}[b!]
    \centering
    \includegraphics[height=0.5\linewidth, width=\linewidth]{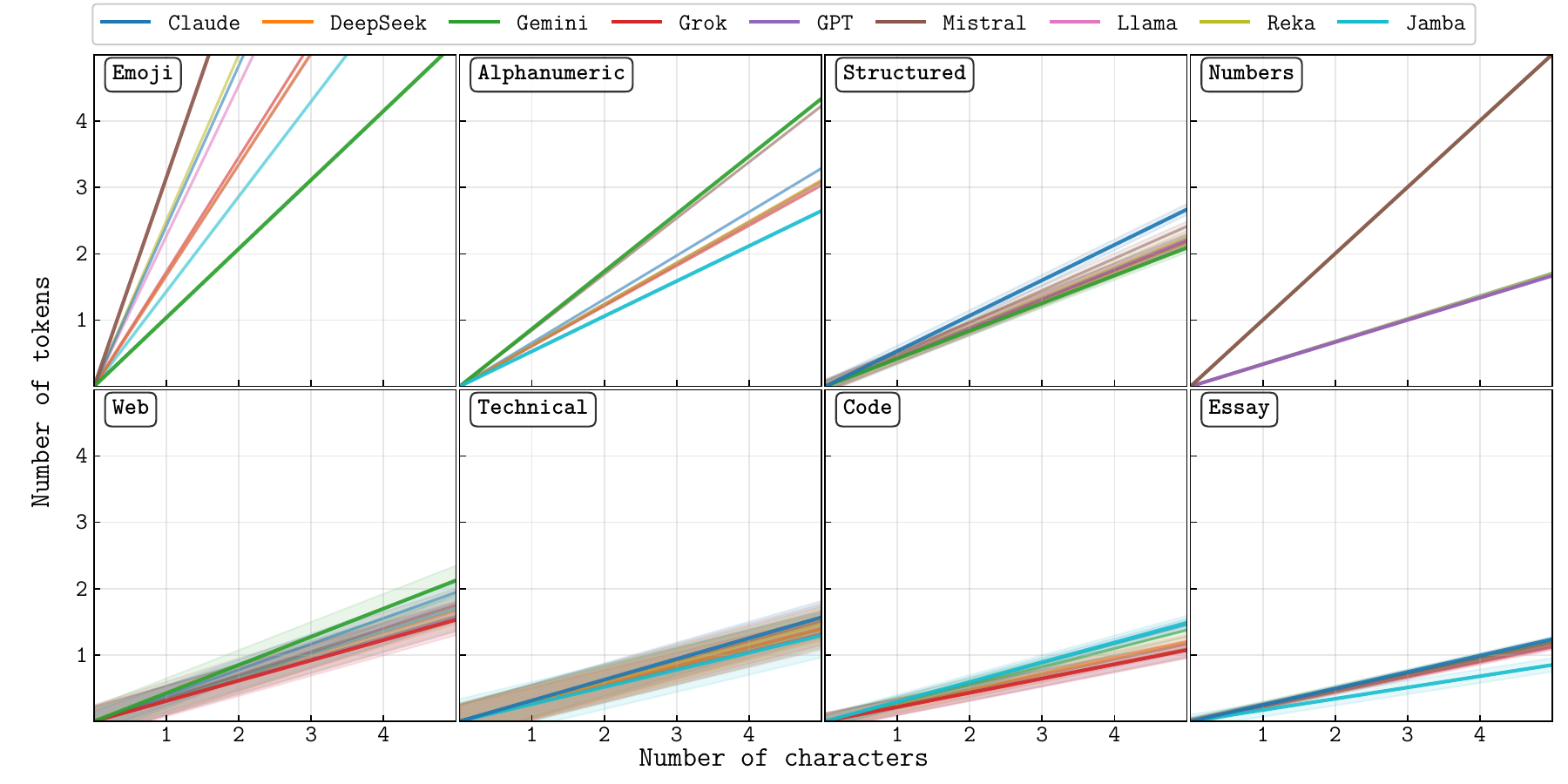}
    \caption{\textbf{Token-character mapping varies significantly across domains and tokenizers}. Lines are plotted using the inverse of the compression ratio as the gradient: steeper slopes require more tokens per character. Shaded regions show 95\% confidence intervals.}
    \label{fig:token_char_mapping}
\end{figure}

A key length metric of LLMs is their \textit{context limit} -- the maximum number of input and output tokens they can process simultaneously -- which is reported in model native tokens (\textit{i.e.}, specific to the model tokenizer). As the compression ratio of text sequences into tokens varies across tokenizers and domains (Fig. \ref{fig:compression_rates}), the ``token'' does not provide a consistent length measurement. This is further evident in the divergent lines of Fig. \ref{fig:token_char_mapping} %
that show mappings derived from the empirical compression ratios (\S\ref{sec:char_comp}). Therefore, using model native token counts is problematic and prevents direct comparisons between models. 

\begin{figure}[t]
    \centering
    \includegraphics[width=\textwidth]{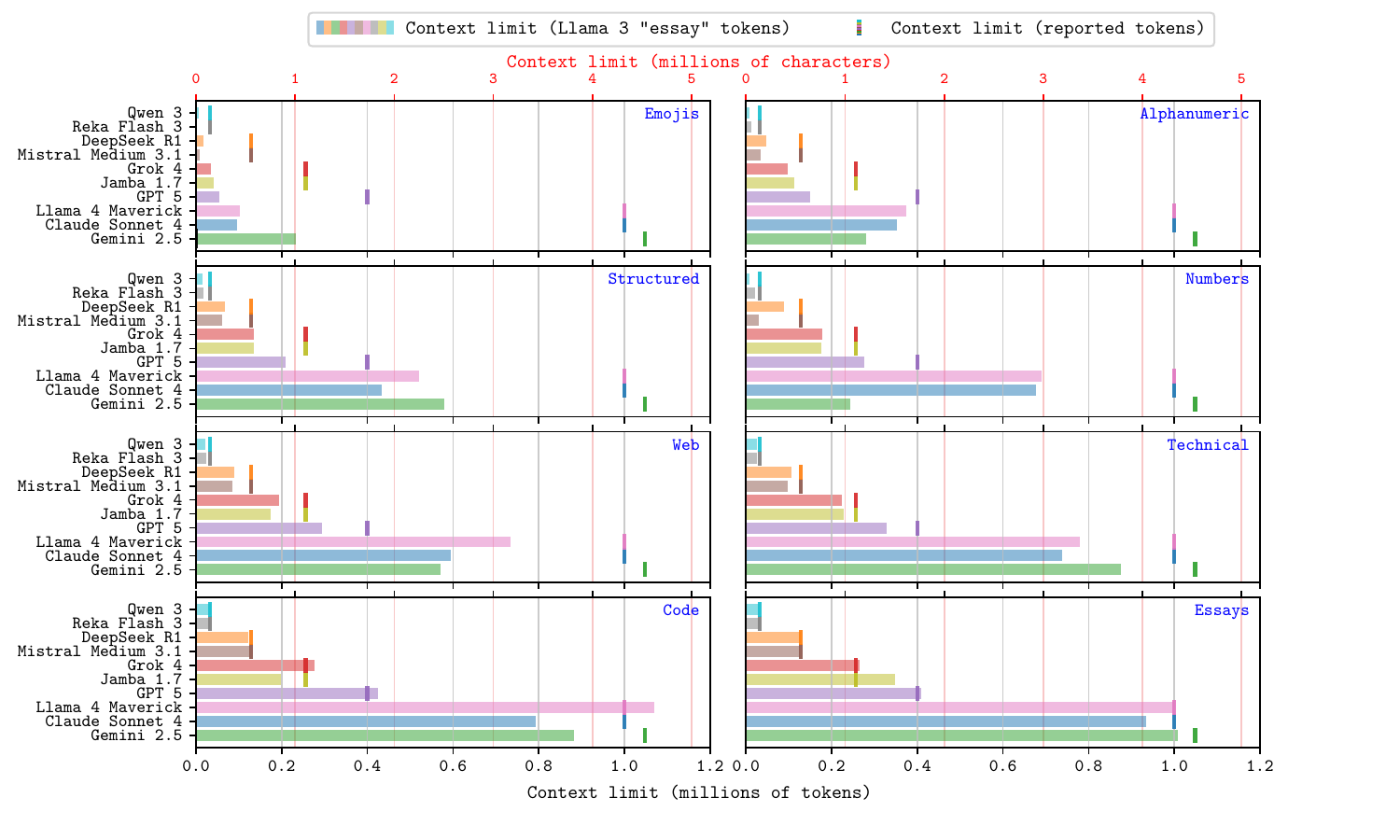}
    \vspace{-0.9cm}
    \caption{\textbf{Comparing context limits}. By converting reported ``model-native'' token context limits into a fixed unit, we can provide direct comparisons of context limits for given text domains. The lower (black) x-axis denotes the context limit in tokens, while the upper (red) axis displays the context limit in characters. Shaded bars represent the context limit in equivalent Llama 3 essay tokens, while the coloured lines denote the reported context limit in model native tokens.}
    \label{fig:context_limits}
\end{figure}

In Fig. \ref{fig:context_limits}, we illustrate the differences in context limits for different models and domains using consistent length representations. Concretely, we initially use the empirical compression ratios and reported model context limits to derive a domain-specific \textit{character context limit} for each model---this value is effectively tokenizer agnostic (see upper x-axis). We then use a set tokenization rate -- specifically, the Llama 3 tokenization rate for natural language (essays) -- to convert the domain-specific character context limits for each model into \textit{equivalent Llama 3 essay token context limits}\footnote{The choice of reference tokenization scheme is arbitrary; we use Llama 3 essay tokenization given the popularity in open source research of Llama and similarity of essay text to ordinary natural language.}. These ``equivalent token'' context limits should be interpreted as a text sequence of equal character length to an equivalent text sequence made up of as many Llama 3 essay tokens. A key observation from this plot is that context limits do not correspond to the same length text sequences across models and domains---even models with equal reported context limits have different length context limits in terms of characters or equivalent tokens. For almost all domains, the equivalent context limit in Llama essay tokens is far below the reported context limit. These results convey an important insight that reported context limits are nuanced: they refer to \textbf{the tokenization of a \textit{specific model} on text of a \textit{specific domain} and should not be directly compared}.

\subsection{Additional experiments}

In the \nameref{app}, we continue our empirical analysis of tokenizers beyond the preceding experiments to investigate compression ratio variation across additional languages and alphabets,  geographic regions, and as text characters are poisoned. We test the hypothesis that compression ratio serves as a crude proxy for prevalence in the LLM training corpus and knowledge.

\section{Related Work}

A number of prior works have carried out empirically-driven analyses of tokenizers, broadly falling along two directions: comparing the effect of (1) differing tokenization ratios on underrepresented languages and (2) tokenizer design on downstream tasks.

\textbf{Language}. A substantial body of research has centered on the differences in tokenization rates across languages. In \cite{tamang2024evaluating}, compression ratios are compared for the 22 official Indian languages. Other studies have found downstream performance to be more robust to the effect of language imbalance in tokenizer training \cite{zhang2022robust}, while compression optimized tokenization segmentation strategies are shown to potentially provide advantages for low-resource language applications \cite{raj2024every}. It is also demonstrated that tokenization methods fail to fairly represent underrepresented complex language scripts \cite{velayuthan2025egalitarian}. Several works compare the impacts of tokenizer choices on downstream multilingual tasks \cite{lotz2025beyond, rust2021good}. Another angle this research has taken is to compare costs: less represented languages in the training corpus often have lower tokenizer compression ratios, typically requiring more input and output tokens \cite{ahia2023all, petrov2023language}. A recent work \cite{turuta2025tokenization} shares similarities with our work as it evaluates contemporary LLM tokenizers, however the focus is on efficiency for  Ukranian language text.

\textbf{Downstream tasks}. Numerous works explore the impacts of tokenizer design on downstream LLM tasks \cite{goldman2024unpacking, saleva2023changes, dagan2024getting, uzan2024greed}. These range from comparing correlations between tokenizer compression ratios and downstream performance \cite{goldman2024unpacking}, the effect of different byte-pair encoding merge operations on performance \cite{saleva2023changes}, to performance metrics such as generation speed and effective context size \cite{dagan2024getting}. Findings from \cite{schmidt2024tokenization}, suggest compression does not provide a clear explanation of what makes a tokenizer effective for downstream tasks.

One recent work \cite{roberts2025needle} has eluded to the disparity between tokenizers of contemporary LLMs, demonstrating that differences in tokenization rates lead to inconsistency in defining sequence lengths. %
However, this analysis only covers the type of abstract alphanumeric text sequences commonly used in needle-in-a-haystack style experiments, thus offers limited widely applicable insights regarding tokenization. Our study provides a much broader and comprehensive evaluation of tokenization rates of commonly used tokenizers across many text domains and frequencies.

\section{Conclusions}

We conduct an empirical analysis of tokenization, focusing on the tokenizers used by frontier LLMs. We curate a small text corpus covering 8 distinct domains and compare the compression ratios of 10 tokenizers, finding significant variation between tokenizers and domains. We report similar variation in the compression ratios of text sequences translated into different languages. Our analysis also challenges the commonly held rule of thumb equating one token to 0.75 words, finding it to be inadequate and oversimplified. We demonstrate the inconsistencies of using tokens as a metric for measuring sequence lengths and comparing model context limits. We hope these empirical insights will lend clarity to the often-overlooked yet crucial process of tokenization in LLMs.

\subsubsection*{Acknowledgments}
This work was supported by the UKRI Centre for Doctoral Training in Application of Artificial Intelligence to the study of Environmental Risks (reference EP/S022961/1), an Isaac Newton Trust grant, a research gift from Google, an EPSRC HPC grant, the Hong Kong Research Grant Council - General Research Fund (Grant No. 17211024), and HKU Seed Fund for PI Research. Samuel would like to acknowledge the support of Z. Novak and N. Novak in enabling his contribution.

\bibliography{iclr2026_conference}
\bibliographystyle{unsrtnat}
\newpage
\FloatBarrier

\appendix
\section*{Appendix}
\label{app}

\section{Additional words per token results}

\begin{figure}[h]
    \centering
    \includegraphics[height=0.5\linewidth, width=\linewidth]{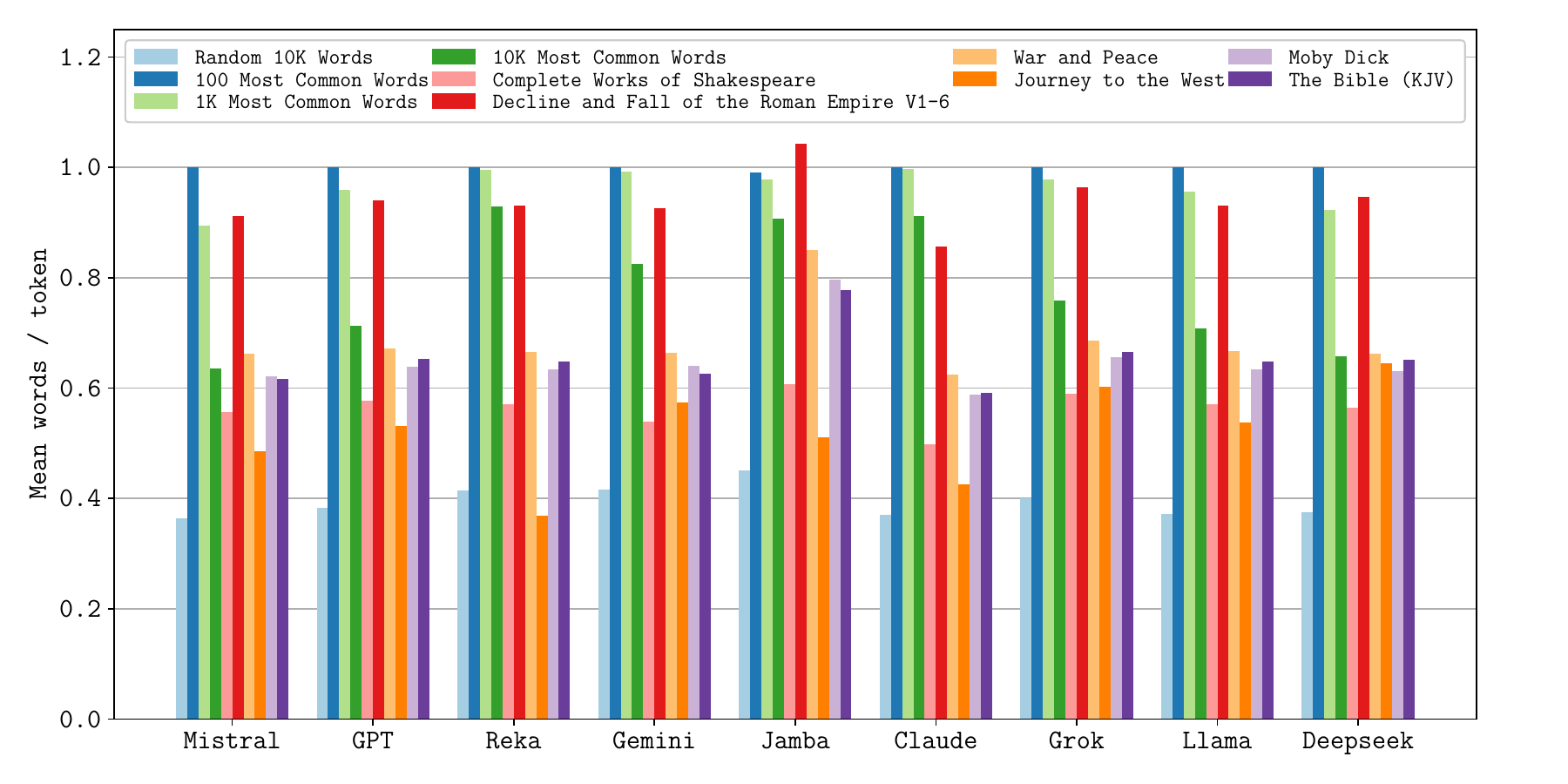}
    \vspace{-0.5cm}
    \caption{\textbf{Average words/token for different text corpora}. Random 10k Words are randomly sampled from WordNet \cite{miller1995wordnet} English words and the 100, 1K, 100K Most Commons Words are derived from the most frequent words in Google's Trillion Word Corpus \cite{google10000english, brants_franz_2006_web1t5}. The books were sourced from Project Gutenberg \cite{ProjectGutenberg}.}
    \label{fig:words_per_token}
\end{figure}

\FloatBarrier

\section{Languages Compression Ratios}

\begin{figure}[h]
    \centering
    \includegraphics[width=\textwidth]{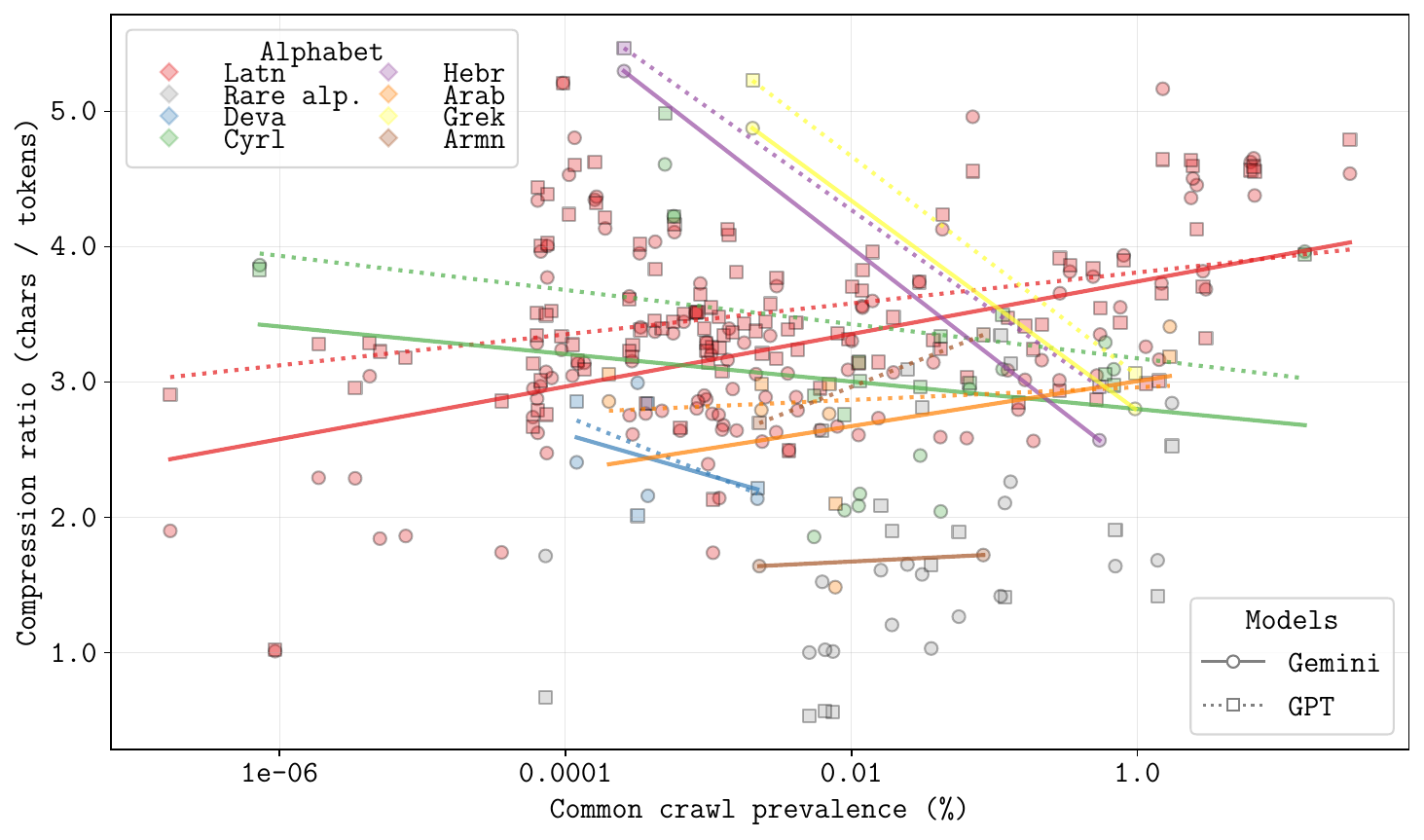}
    \vspace{-0.5cm}
    \caption{\textbf{Compression ratios for a natural language essay translated into languages present in Common Crawl}. Each data point represents a language text tokenized by the Gemini (circle) or GPT (square) tokenizers. Language common crawl prevalence is estimated from \cite{abadji-etal-2022-towards}. Translated data was derived using GPT-4o to translate the essay ``What to do" \cite{graham_essays_web}.}
    \label{fig:alphabet_tokenization}
\end{figure}

\FloatBarrier

\section{Geospatial}

\begin{figure}[h]
    \centering
    \includegraphics[width=\textwidth]{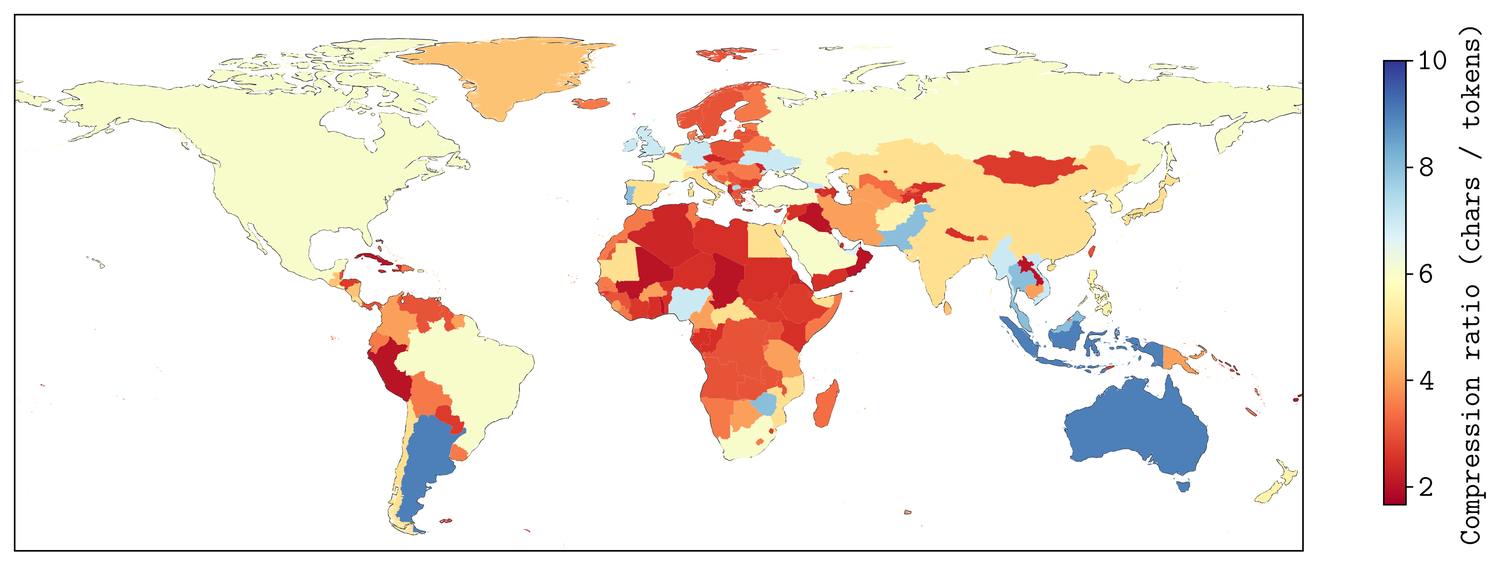}
    \vspace{-0.5cm}
    \caption{\textbf{Country name tokenization}. Compression ratios are displayed for the GPT tokenizer using country names and geometries derived from cartopy \cite{cartopy}.}
    \label{fig:country_tokens}
\end{figure}

\begin{figure}[h]
    \centering
    \includegraphics[width=\textwidth]{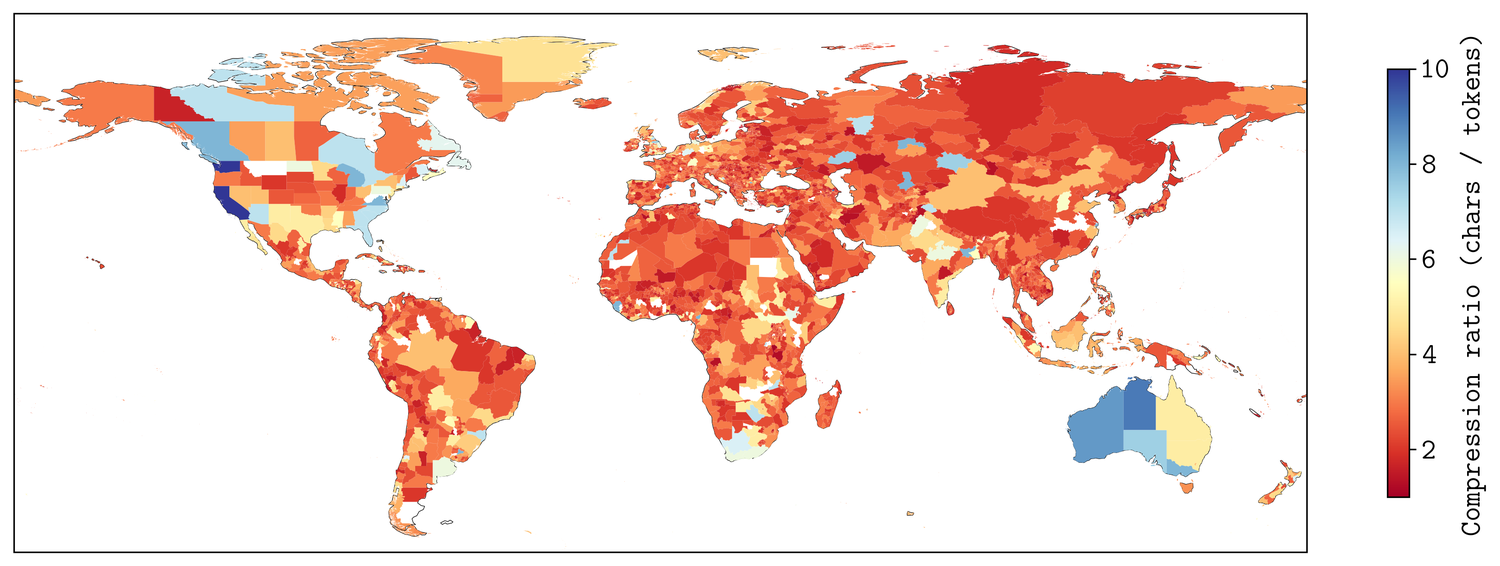}
    \vspace{-0.5cm}
    \caption{\textbf{Region name tokenization}. Compression ratios are displayed for the GPT tokenizer using region names and geometries derived from cartopy \cite{cartopy}.}
    \label{fig:region_tokens}
\end{figure}

\begin{figure}[h]
    \centering
    \includegraphics[width=\textwidth]{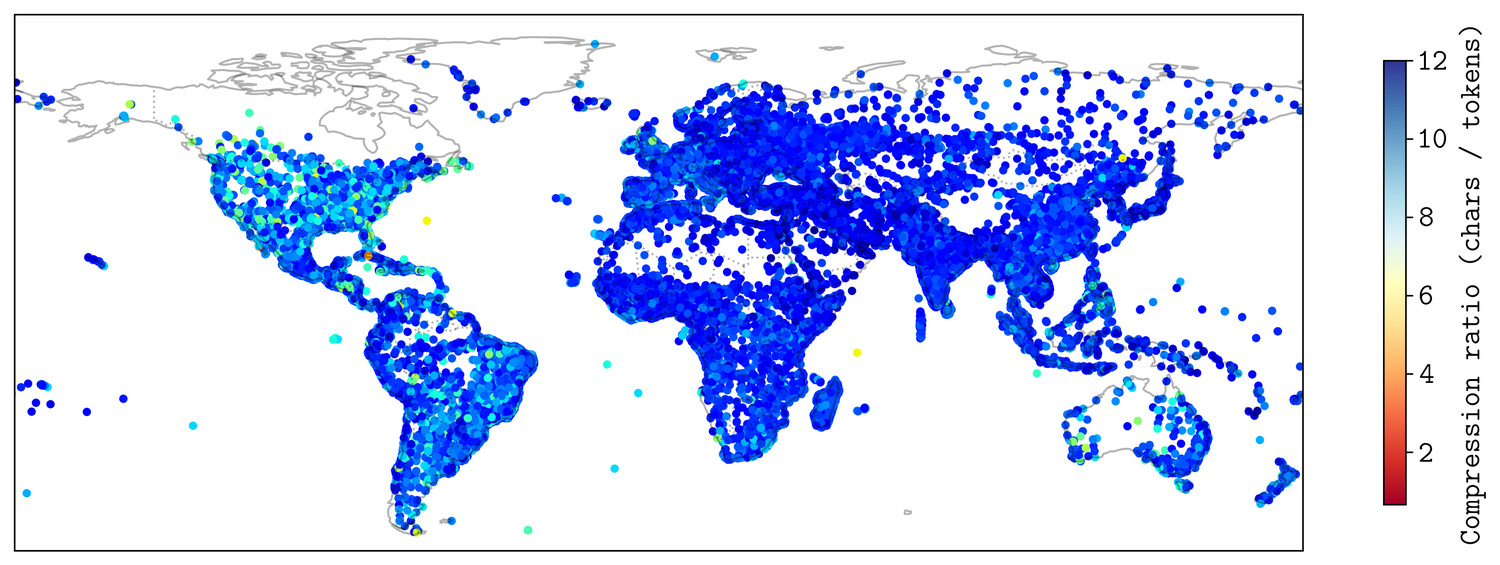}
    \vspace{-0.5cm}
    \caption{\textbf{Settlement name tokenization}. Compression ratios are displayed for the GPT tokenizer for $\sim$48k town and city names and geometries from \cite{simplemaps_com_2025}.}
    \label{fig:city_tokens}
\end{figure}

The geospatial experiments in this subsection investigate hypotheses related to potential geographic bias in the tokenizer training corpora. Figs. \ref{fig:country_tokens}-\ref{fig:city_tokens} display the GPT tokenizer compression ratios for country, region and city/town names, respectively. 

\begin{figure}[h]
    \centering
    \includegraphics[width=\textwidth]{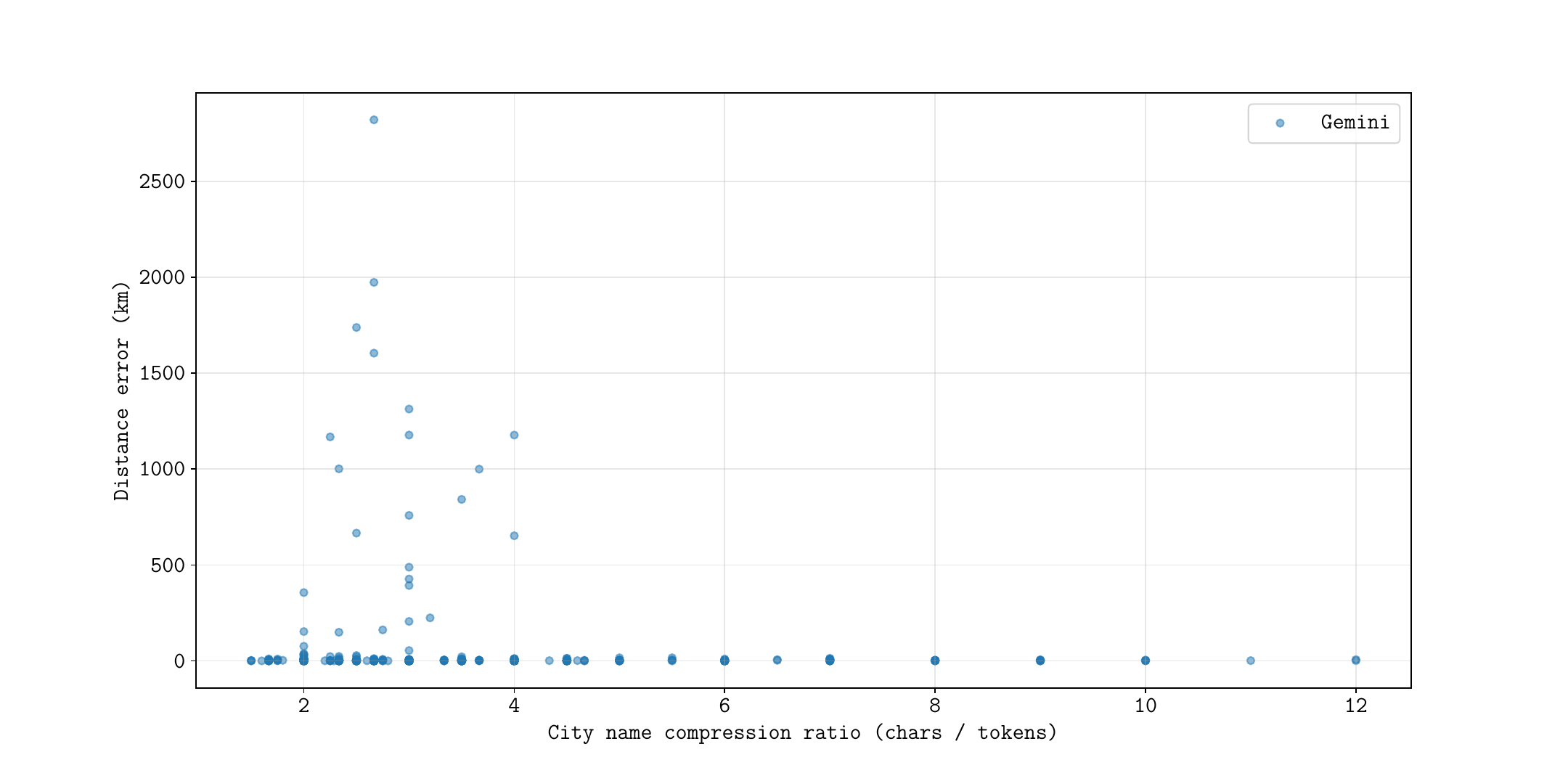}
    \vspace{-0.5cm}
    \caption{\textbf{Haversine distance errors} between the positions (latitude, longitude) predicted by Gemini 2.5 Flash and ground truth for 48k settlements worldwide \cite{simplemaps_com_2025}.}
    \label{fig:city_distance_err}
\end{figure}

Fig. \ref{fig:city_distance_err} shows the haversine distance error between the latitude and longitude positions predicted by Gemini 2.5 Flash and ground truth of 48k settlements worldwide (see Fig. \ref{fig:city_tokens}). When plotted against the compression ratios of the settlement names, a negative correlation is observable: distance error decreases with higher compression ratios. We hypothesize this is tied to the distribution of training data. Assuming the tokenizer training data follows a similar distribution to the LLM training data, words or subwords that are more prevalent will both have a higher compression and represent concepts the LLM has more accurate knowledge of.

\FloatBarrier

\section{Character Poisoning}

\begin{figure}[h]
    \centering
    \includegraphics[width=0.7\textwidth]{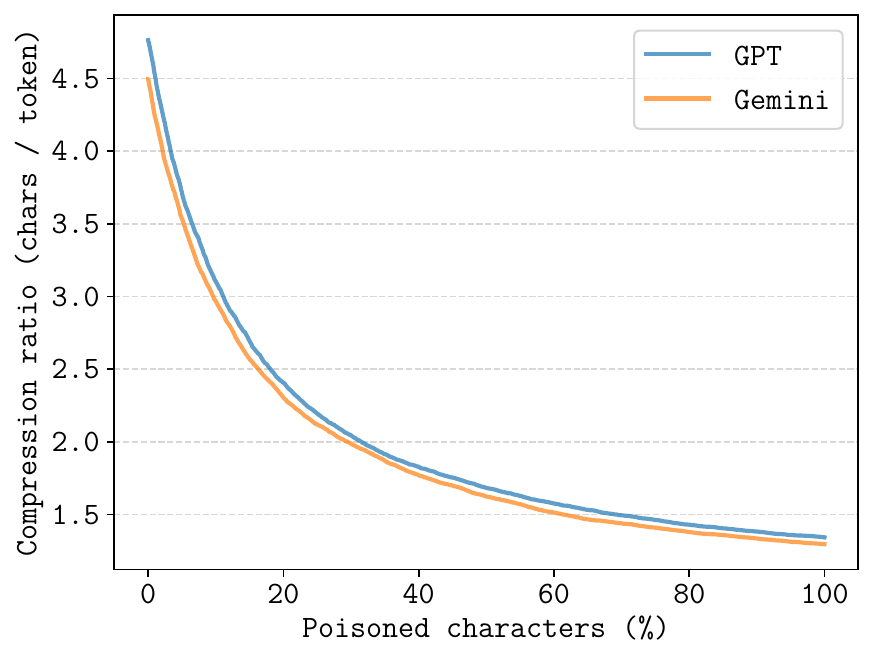}
    \caption{\textbf{Compression ratios decay exponentially as characters are randomly poisoned} for text from ``What to do" \cite{graham_essays_web}. Incrementally, characters were selected at random and replaced with a different character. The compression ratio was averaged across the entire text each iteration.}
    \label{fig:token_poisoning}
\end{figure}

\end{document}